\AtBeginDocument{%
  }
\documentclass{svproc}
\usepackage{graphicx}
\usepackage{graphics}
\usepackage{url}
\usepackage{multirow}
\renewcommand{\arraystretch}{1.3}
\usepackage{booktabs}
\usepackage{tabularx}
\usepackage{url}
\usepackage{amssymb}
\renewcommand{\arraystretch}{1.2}
\begin{document}

\mainmatter              % start of a contribution
\title{Statistical Machine Translation Systems of English-Pnar Language Pair : Some Insights of the Emperical Study}
\titlerunning{Statistical Machine Translation Systems of English-Pnar..}  % abbreviated title (for running head)
%                                     also used for the TOC unless
%                                     \toctitle is used
%
\author{
Edawanbiang Dhar\inst{1} \and
Surmila Thokchom\inst{2} \and
Thoudam Doren Singh\inst{1} 
}

\authorrunning{Edawanbiang Dhar et al.}

\tocauthor{
Edawanbiang Dhar, Surmila Thokchom Thoudam Doren Singh   
}

\institute{
Human Language Technology Lab, Department of Computer Science and Engineering,\\
National Institute of Technology Meghalaya, Sohra, Meghalaya-793108, India\\
\email{p24cs005@nitm.ac.in, doren.thoudam@nitm.ac.in}
\and
Department of Computer Science and Engineering,\\
National Institute of Technology Meghalaya, Sohra, Meghalaya-793108, India\\
\email{surmila.thokchom@nitm.ac.in}
}

\maketitle              % typeset the title of the contribution

\begin{abstract}

  Pnar, an Austroasiatic language spoken by approximately 0.4 million people in the Jaintia Hills of Meghalaya, lacks the digital corpora and natural language processing (NLP) resources. This paper presents the first machine translation study for the English--Pnar language pair. Using articles collected from the \textit{Wyrta} newspaper, we built a parallel corpus comprising of 10,234 sentences and trained phrase-based statistical machine translation (SMT) systems the models using 9,563 parallel corpora under three configurations for each direction using Moses, GIZA++, KenLM, varying lexicalized reordering and minimum error rate training (MERT) tuning. The models are evaluated on a held-out test set of 371 sentences, the best-performing system achieves a BLEU score of 14.97 (chrF2: 33.42, TER: 77.60) for Pnar$\rightarrow$English and 11.16 (chrF2: 31.38, TER: 93.51) for English$\rightarrow$Pnar, establishing the first quantitative benchmark for this language pair. Lexicalized reordering improves translation quality by 3.73 BLEU points for Pnar$\rightarrow$English, reflecting the structural shift from the source language's SOV word order to the target language's SVO order, whereas MERT tuning degrades BLEU performance under low-resource conditions. Finally, we analyze the remaining translation errors, including morphological out-of-vocabulary (OOV) words, long-distance reordering, and Khasi code-mixing and discuss future directions toward neural and multilingual machine translation for Pnar.

\end{abstract}
\section{Introduction}
Pnar is a severely under-resourced language spoken by the Jaintia community in Meghalaya in the Northeast India. Despite its linguistic proximity to Khasi which has received limited attention in natural language processing, Pnar remains largely unexplored in computational research. It is written in the Roman script and belongs to the Austroasiatic language family characterized by distinct morphological and syntactic features that distinguish it from more widely studied Indo-European languages. The scarcity of computational resources for Pnar reflects broader challenges faced by low-resource languages in natural language processing \cite{araabi2020optimizing}. Developing language technologies for such languages is crucial for their preservation and continued use and particularly by enabling access to digital content, education and communication tools for native speakers \cite{bird2020decolonising}. In this context, establishing a Pnar-English machine translation system represents an important step toward advancing research in low-resource NLP as well as supporting linguistic sustainability. Pnar, also known as Jaintia (Spencer 1967) or earlier as Synteng (Grierson 1928), is a language spoken in the East and West Jaintia Hill District of Meghalaya and in a few pockets of Cachar Hills and North Cachar Hills districts of Assam \cite{Bareh2026}. According to the People’s Linguistic Survey of India (2014), the population of Pnar stands at 3,92,853. Pnar is also subject to regional variations which can be noticeable even within a distance of 3-4 kilometres. The most noticeable variants of Pnar can be classified according to their regional locations; in most cases, the regional variants take their names from village names \cite{Bareh2026}. To address these challenges, we document this low-resource language, Pnar, by building a corpus using local newspapers (Wyrta). Machine Translation (MT) has witnessed remarkable progress in recent years, driven by large-scale parallel corpora and powerful neural architectures such as the Transformer \cite{vaswani2017attention}.  The vast majority of the world's approximately 7,000 languages remain severely under-resourced, lacking both parallel data and computational tools necessary for developing effective MT systems \cite{araabi2020optimizing}. This paper makes three main contributions. First, we built the first sizeable parallel corpus for Pnar of a local newspaper. Second, we train and compare phrase-based statistical machine translation (SMT) systems under six configurations (three per translation direction), quantifying the individual effects of lexicalized reordering and minimum error rate training (MERT) tuning in low-resource settings. Third, we establish the first published baseline for English--Pnar machine translation using BLEU, chrF2, and TER evaluation metrics, together with an error analysis that highlights challenges arising from Pnar morphology and its code-mixing with Khasi. Unlike prior machine translation studies on other Northeast Indian languages, such as Manipuri, Mizo, Bodo, and Khasi, no computational study has previously targeted Pnar. This work therefore addresses an important gap in the literature.

\section{Related Work} Low-resource machine translation has attracted considerable research attention. Araabi and Monz~\cite{araabi2020optimizing} showed that Transformer architecture optimization, including careful regularization and reduced model capacity benefits low-resource neural MT settings. Bird~\cite{bird2020decolonising} discusses the broader decolonization imperative in speech and language technology arguing that community-centered approaches are essential for endangered language documentation. For Northeast Indian languages, SMT and neural MT systems have been reported for Manipuri, Mizo, Bodo, and Khasi. These prior efforts share a common approach namely, phrase-based or low-resource neural machine translation (MT) systems trained on small, domain-specific parallel corpora. However, none has been extended to Pnar, whose closest documented linguistic relative, Khasi, has only recently begun to receive attention from the computational linguistics community. This gap is significant because the linguistic characteristics of Pnar including its SOV constituent order, agglutinative verbal morphology and frequent code-mixing with Khasi, make it substantially different from previously studied low-resource languages. In addition to these structural differences from English, Pnar also suffers from an acute scarcity of digital resources, including even monolingual text. Consequently, the present study is not merely a replication of existing SMT pipelines developed for Khasi or Manipuri; rather, it represents the first systematic investigation of how effectively a standard phrase-based SMT framework performs under these combined linguistic and resource constraints while identifying the specific challenges that limit its performance. Among Austroasiatic languages, there is also report on the initiative of developing Santali--English translation system. However, to the best of our knowledge, no prior computational MT work has addressed Pnar specifically making this study the first of its kind. The Moses toolkit~\cite{koehn-etal-2007-moses} used in this study has been widely applied in low-resource scenarios due to its robustness with small training sets and its well-understood behavior. KenLM~\cite{heafield-2011-kenlm} provides an efficient language model backend that makes higher-order $n$-gram models practical even in resource-constrained environments. MGIZA++~\cite{gao-vogel-2008-parallel} enables parallel word alignment using IBM Models substantially reducing training time.

\section{Dataset Preparation}
\subsection{Data Sources}
Given the scarcity of digital Pnar text, corpus construction required identification and digitization from multiple sources. We collected Pnar dataset from a single primary source. First, contents of the Wyrta newspaper are collected, providing contemporaneous general-domain text covering community news, local governance, cultural events, sports and health topics. 

\begin{table}[htbp]
\centering
\caption{Sentence counts, token counts and average sentence length for the train, development and test splits of the English-Pnar parallel corpus statistics}
\label{tab:corpus}

\footnotesize
\renewcommand{\arraystretch}{1.1}

\begin{tabular}{p{2.8cm} p{1.6cm} p{1.6cm} p{1.6cm}}
\toprule
\textbf{Metric} & \textbf{Train} & \textbf{Dev} & \textbf{Test} \\
\midrule
Sentences        & 9,563 & 300 & 371 \\
Pnar tokens      & $\sim$234,800 & $\sim$7,350 & $\sim$5,777 \\
English tokens   & $\sim$229,400 & $\sim$7,200 & $\sim$5,865 \\
Average Pnar sentence length   & $\sim$24.6 & $\sim$24.5 & $\sim$15.6 \\
Average English sentence length   & $\sim$24.0 & $\sim$24.0 & $\sim$15.9 \\
\bottomrule
\end{tabular}
\end{table}

\subsection{Data Preprocessing} 
Data preprocessing follows standard SMT pipeline conventions. Text is lowercased and tokenised using the Moses tokeniser adapted for Roman-script Pnar text. Punctuation normalisation is applied to handle inconsistencies across newspaper scans and OCR output. Since Pnar lacks a standard spell-checker, manual lexicon-based correction was performed for the 500 most frequent tokens. Sentence pairs with either side exceeding 80 tokens were removed to improve alignment quality. Trailing comma artefacts introduced during corpus splitting are cleaned using regular expression filters. The final cleaned corpus is split into training (9,563 sentences), development (300 sentences, lines 9,563--9,863, strictly non overlapping with training), and test sets (371 sentences from a held-out evaluation set).

\section{Experimental Setup}
\label{experimental setup}
A phrase-based model is trained using the Moses toolkit \cite{koehn-etal-2007-moses}. Moses is a widely used open-source framework that implements phrase-based translation models, supporting modular components such as translation models, reordering models and decoding strategies. The SMT pipeline includes word alignment using GIZA++, phrase extraction and a 5-gram language model trained with KenLM \cite{heafield-2011-kenlm} on the target-side training data. Standard Minimum Error Rate Training (MERT) tuning is applied on the validation set to finetune translation quality. Fig 1. shows the Phrase-based SMT model architecture of the proposed translation system.

\subsubsection{Moses SMT Framework}
We train phrase-based Statistical Machine Translation (SMT) systems using the Moses toolkit \cite{koehn-etal-2007-moses}, a widely used open-source framework that implements phrase-based translation models with modular components, including translation models, reordering models, and decoding strategies. The SMT approach formulates translation as the problem of finding the most probable target sentence $e^*$ given a source sentence $f$, defined as \ref{eq:smt}:
\begin{equation}
e^* = \arg\max_{e} \; p(f \mid e)\, p(e)
\label{eq:smt}
\end{equation}

where $p(f \mid e)$ represents the translation model probability and $p(e)$ denotes the language model probability. The decoder performs a search for the optimal hypothesis using beam search over the combined model score.

\begin{figure}[ht]
  \centering
  \includegraphics[width=0.5\textwidth]{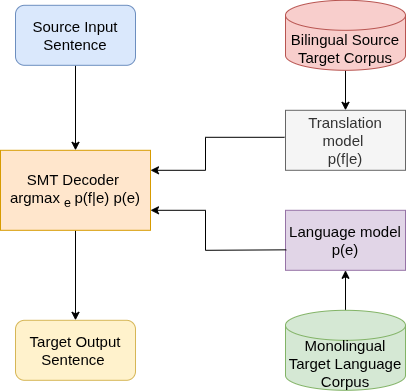}
  \caption{Phrase-based SMT model architecture : the decoder combines translation-model and language-model scores over the bilingual and monolingual training corpora to produce the target sentence.}
  \label{fig:smt}
\end{figure}

\subsection{Word Alignment and Phrase Extraction}
Phrase alignment is performed using GIZA++~\cite{och-ney-2003-systematic}  implementing IBM Models in both source-to-target and target-to-source directions. The resulting bidirectional alignments are symmetrized using the \textit{grow-diag-final-and} heuristic~\cite{koehn-etal-2003-statistical}, which has shown to produce high-quality phrase alignment and improved phrase coverage. Phrase pairs are extracted from the symmetrized word alignments with a maximum phrase length of 7 words. Phrase translation probabilities are estimated using relative frequency over extracted phrase pair counts. Lexical weighting features are computed from word-level translation tables to provide more fine-grained translation scoring. Having established the corpus and evaluation protocol (Section~\ref{experimental setup}), we now describe the three main components of the statistical machine translation (SMT) pipeline, word alignment, the language model and the reordering model followed by the tuning procedure that integrates them.

\subsection{Language Model}
For Pnar$\to$English systems, the language model is trained on the 8,05,946 English target-side training sentences. For English$\to$Pnar systems, the language model is trained on a monolingual Pnar corpus of 23,429 sentences drawn from \textit{Wyrta} newspaper issues. Two 5-gram KenLM language models are built on the target-side data using the 8,05,946 English target monolingual sentences for Pnar$\rightarrow$English translation and a 23,429-sentence monolingual Pnar corpus for English$\rightarrow$Pnar translation respectively. In both cases, a 5-gram language model with modified Kneser--Ney smoothing is trained using KenLM~\cite{heafield-2011-kenlm} which utilises probing hash tables and trie-based data structures to achieve fast query performance and reduced memory usage. The language model plays a critical role in ensuring fluent and grammatically coherent translations by assigning higher probabilities to well-formed target sentences as given by equation \ref{eq:lm} :

\begin{equation}
p(e) = \prod_{i=1}^{n} p(e_i \mid e_1^{i-1})
\label{eq:lm}
\end{equation}

\subsection{Reordering Model}
\label{reordering}
Given the SOV (Subject--Object--Verb) versus SVO (Subject--Verb--Object) word order divergence between Pnar and English, lexicalized reordering models play a crucial role in improving translation quality for this language pair. We employ a lexicalized reordering model based on phrase-pair orientation types including \textit{monotone}, \textit{swap}, \textit{discontinuous-left} and \textit{discontinuous-right} configurations ~\cite{tillmann2004unigram}. Reordering model weights are estimated from the training data jointly with phrase translation probabilities within the phrase-based SMT framework~\cite{koehn-etal-2007-moses}. This component is particularly important for handling Pnar verb-final constructions which require reordering when translated into English as well as for managing differences in nominal modifier ordering between the two languages.

\subsection{Tuning} 
Minimum Error Rate Training (MERT) ~\cite{och2003minimum} is applied on the validation set to optimize the log-linear combination of model feature weights with respect to the BLEU score ~\cite{papineni2002bleu}. The SMT system is modeled as a log-linear combination of features, including phrase translation probabilities, language model scores, reordering model features, and a word penalty term. The feature weights are tuned on the development set using MERT for up to 25 iterations or until convergence is reached. This optimization procedure directly maximizes translation quality as measured by automatic evaluation metrics such as BLEU. Three system configurations are trained and evaluated in each translation direction (Table~\ref{tab:configs}).

\begin{table}[htbp]
\centering
\caption{The three SMT configurations in each translation direction varying lexicalized reordering and MERT tuning}
\label{tab:configs}

\footnotesize
\renewcommand{\arraystretch}{1.1}

\begin{tabular}{p{2.0cm} p{3.2cm} p{2.6cm} p{2.6cm}}
\toprule
\textbf{System} & \textbf{Reordering} & \textbf{Tuning} & \textbf{LM} \\
\midrule
Config-1 &
$\checkmark$ msd-bidirectional &
None (untuned) &
5-gram KenLM \\

Config-2 & $\checkmark$ msd-bidirectional &
MERT (25 iter.) &
5-gram KenLM \\

Config-3 &
$\times$ None &
None (untuned) &
5-gram KenLM \\
\bottomrule
\end{tabular}
\end{table}

\section{Experiments and Results}

\subsection{Results} Tables~\ref{tab:translation_results}(a) and~\ref{tab:translation_results}(b) present evaluation results
for all configurations on the Pnar$\to$English and
English$\to$Pnar test sets respectively.

\begin{table}[!htbp]
\centering
\caption{Translation performance on the 371-sentence test set.}
\label{tab:translation_results}

\begin{minipage}[t]{0.48\textwidth}
\centering
\textbf{(a) Pnar $\rightarrow$ English}

\vspace{0.3em}

\small
\renewcommand{\arraystretch}{1.2}
\begin{tabular}{lccc}
\toprule
Configuration & BLEU $\uparrow$ & chrF2 $\uparrow$ & TER $\downarrow$ \\
\midrule
Config-3 & 11.24 & 27.44 & 91.70 \\
Config-1 & \textbf{14.97} & 33.42 & \textbf{77.60} \\
Config-2 & 12.43 & \textbf{36.86} & 95.04 \\
\bottomrule
\end{tabular}
\end{minipage}
\hfill
\begin{minipage}[t]{0.48\textwidth}
\centering
\textbf{(b) English $\rightarrow$ Pnar}

\vspace{0.3em}

\small
\renewcommand{\arraystretch}{1.2}
\begin{tabular}{lccc}
\toprule
Configuration & BLEU $\uparrow$ & chrF2 $\uparrow$ & TER$\downarrow$ \\
\midrule
Config-3 & \textbf{11.16} & 31.38 & \textbf{93.51} \\
Config-1 & 10.14 & 30.98 & 86.53 \\
Config-2 & 6.91 & \textbf{32.82} & 124.93 \\
\bottomrule
\end{tabular}
\end{minipage}

\end{table}

\subsection{Impact of Lexicalized Reordering}

Table~\ref{tab:reorder} summarises the impact of the lexicalized reordering model on translation performance.

\begin{table}[htbp]
\centering
\caption{BLEU, chrF, TER scores with vs. without lexicalized reordering, isolating the reordering effect from MERT tuning (both rows untuned).}
\label{tab:reorder}

\small
\renewcommand{\arraystretch}{1.1}

\begin{tabular}{llccc}
\toprule
\textbf{Direction} & \textbf{Config} & \textbf{BLEU $\uparrow$} & \textbf{chrF2$\uparrow$} & \textbf{TER$\downarrow$} \\
\midrule

\multirow{3}{*}{Pn$\to$En}
& No reorder   & 11.24 & 27.44 & 91.70 \\
& Reorder      & \textbf{14.97} & \textbf{33.42} & \textbf{77.60} \\
& $\Delta$     & +3.73 & +5.98 & -14.10 \\
\midrule

\multirow{3}{*}{En$\to$Pn}
& No reorder   & \textbf{11.16} & \textbf{31.38} & 93.51 \\
& Reorder      & 10.14 & 30.98 & \textbf{86.53} \\
& $\Delta$     & -1.02 & -0.40 & -6.98 \\
\bottomrule

\end{tabular}
\end{table}
The reordering model provides a substantial gain of $+3.73$ BLEU and $+5.98$ chrF2 for Pnar$\to$English reflecting the significant SOV$\to$SVO structural transformation required in this direction. The TER improvement of $-14.10$ is particularly noteworthy indicating that reordered hypotheses require substantially fewer post-editing operations. For English$\to$Pnar, the reordering model produces a marginal BLEU decline ($-1.02$). Qualitative analysis reveals that without reordering the En$\to$Pn model largely copies source English tokens with minor shuffling yielding artificially inflated BLEU scores through named entity. The with-reordering system (Config-1, BLEU~10.14) produces genuine Pnar output and constitutes the better baseline. This asymmetry where lexicalized reordering improves Pnar$\rightarrow$English translation but not the reverse direction is consistent with the behavior of reordering models under low-resource conditions. Specifically, the model is able to learn the transformation from the Pnar SOV word order to the English SVO structure thereby reducing structural divergence during decoding. In contrast, the reverse transformation from English SVO to Pnar SOV is more difficult to capture using the same limited phrase table as English provides comparatively fewer local reordering cues for generating the target-side Pnar word order. Consequently, the benefits of lexicalized reordering are more pronounced for Pnar$\rightarrow$English translation than for English$\rightarrow$Pnar translation.

\subsection{Effect of MERT Tuning}
\label{MERT}
MERT tuning (Config-2) degrades BLEU in both translation directions ($-2.50$ for Pnar$\to$English and $-3.23$ for English$\to$Pnar) while improving chrF2 marginally. The tuned models exhibit output length ratios exceeding 1.0 (over-generation) suggesting the MERT weight optimisation converges to degenerate solutions under the small corpus setting. This is consistent with known limitations of MERT under low-resource conditions~\cite{araabi2020optimizing}, where the development set is insufficient to reliably estimate 14 feature weights. Despite degraded BLEU, the higher chrF2 for MERT-tuned systems suggests better character-level coverage, warranting further investigation. The chrF2/BLEU divergence under MERT tuning is itself informative: chrF2 rewards character-n-gram overlap and is comparatively robust to the length and word-order errors that BLEU penalizes heavily, so an over-generating system can gain chrF2 while losing BLEU. This suggests that MERT, optimizing directly against BLEU on a 300-sentence dev set with 14 free feature weights, overfits to length statistics of that small set rather than to genuine translation adequacy. In practice, this means BLEU-tuned weights are not a reliable model-selection criterion at this corpus size; a held-out chrF2 or human-adequacy check would be a safer stopping criterion for future low-resource SMT work.

\subsection{Statistical Significance}
\label{sec:significance}

To assess whether the differences reported in Table~\ref{tab:translation_results} reflect
genuine system differences rather than test-set sampling variance, we apply
paired bootstrap resampling~\cite{koehn2005edinburgh} with 1{,}000 resamples
over the full 371-sentence test set in each translation direction, using
SacreBLEU (detokenized, \texttt{tok:13a}) to compute corpus-level BLEU on
each resample. For Pnar$\rightarrow$English, the reordering gain (Config-1 vs.\ Config-3) is
$+3.72$ BLEU (95\% CI $[3.16, 4.30]$, $p < 0.001$), confirming that
lexicalized reordering yields a genuine, statistically significant
improvement rather than an artefact of test-set sampling. MERT tuning
(Config-2 vs.\ Config-1) produces a significant \emph{decrease} of $-2.50$
BLEU (95\% CI $[-3.24, -1.75]$, $p < 0.001$), confirming that the
degradation reported in Section~\ref{MERT} is a robust effect rather
than sampling noise.

For English$\rightarrow$Pnar, the small BLEU decline attributed to
reordering (Config-1 vs.\ Config-3, $-1.02$ BLEU) is \emph{not} statistically
significant (95\% CI $[-2.19, 0.11]$, $p = 0.084$): the confidence interval
crosses zero, so we cannot reject the possibility of no true difference at
this test-set size. This supports the qualitative explanation in
Section~\ref{reordering} that the apparent advantage of the no-reorder
system is likely an artefact of named-entity copying rather than a genuine
translation-quality difference. In contrast, MERT tuning's degradation
(Config-2 vs.\ Config-1, $-3.22$ BLEU) is highly significant (95\% CI
$[-3.96, -2.43]$, $p < 0.001$), showing that MERT's harm under this
low-resource setting is a consistent effect across both translation
directions.

\begin{table}[t]
\centering
\caption{Paired bootstrap significance test ($B{=}1{,}000$) on the
371-sentence test set. $\dagger$ = significant at $p<0.05$.}
\label{tab:significance}
\begin{tabular}{llrrrc}
\toprule
Direction & Comparison & $\Delta$ BLEU & 95\% CI & $p$-value & Sig.? \\
\midrule
Pn$\rightarrow$En & Config-1 vs.\ Config-3 (reordering) & $+3.72$ & $[3.16,\ 4.30]$   & $<0.001$ & Yes$^\dagger$ \\
Pn$\rightarrow$En & Config-1 vs.\ Config-2 (MERT)        & $-2.50$ & $[-3.24,\ -1.75]$ & $<0.001$ & Yes$^\dagger$ \\
En$\rightarrow$Pn & Config-1 vs.\ Config-3 (reordering)  & $-1.02$ & $[-2.19,\ 0.11]$  & $0.084$  & No \\
En$\rightarrow$Pn & Config-1 vs.\ Config-2 (MERT)        & $-3.22$ & $[-3.96,\ -2.43]$ & $<0.001$ & Yes$^\dagger$ \\
\bottomrule
\end{tabular}
\end{table}

\subsection{Sample Translations and Error Analysis}
\label{Sample Translations and Error Analysis}
Tables~\ref{tab:sample_pn2en} and~\ref{tab:sample_en2pn} present
representative translation examples from the best-performing
configurations.

\begin{table*}[!htbp]
\centering
\caption{Sample translations from best Pnar$\to$English system (Config-1, BLEU 14.97). System output is from the untuned, reordering-enabled configuration (Config-1); word-for-word artefacts are discussed in \ref{Sample Translations and Error Analysis}}
\label{tab:sample_pn2en}

\begin{tabularx}{\textwidth}{p{2.2cm}X}
\toprule
\textbf{Type} & \textbf{Sentence} \\
\midrule

Source 1 &
Sdang ka JAC yei yakhih wow khang yei lai i wan ki Truck ka Star Cement wa Dalmia. \\

Reference &
The JAC began a protest to stop the movement of trucks belonging to Star Cement and Dalmia Cement. \\

System &
The JAC yakhih to the khang the on the wan Truck Star Cement along with the Dalmia. \\
\midrule

Source 2 &
Mynsoo joor ki ar ngut hadien wa noh ka Truck chapoh riat ha Ri-Bhoi. \\

Reference &
Two people were seriously injured after a truck fell into a gorge in Ri-Bhoi. \\

System &
Two the ar ngut after the noh Truck chapoh riat in Ri-Bhoi. \\
\midrule

Source 3 &
Daw pynman ka WJDBA ya ka Inter School Badminton Competition 2026. \\

Reference &
WJDBA will organize the Inter-School Badminton Competition 2026. \\

System &
The WJDBA of the Inter School Badminton Competition 2026. \\
\bottomrule
\end{tabularx}
\end{table*}

\begin{table*}[!htbp]
\centering
\caption{Sample translations from best English$\to$Pnar system (Config-1, BLEU 10.14).System output is from the untuned, reordering-enabled configuration (Config-1); word-for-word artefacts are discussed in \ref{Sample Translations and Error Analysis}}
\label{tab:sample_en2pn}

\begin{tabularx}{\textwidth}{p{2.2cm}X}
\toprule
\textbf{Type} & \textbf{Sentence} \\
\midrule

Source 1 &
The JAC began a protest to stop the movement of trucks belonging to Star Cement and Dalmia Cement. \\

Reference &
Sdang ka JAC yei yakhih wow khang yei lai i wan ki Truck ka Star Cement wa Dalmia. \\

System &
Ya ka JAC began protest toh ka movement trucks belonging Star Cement wa u Dalmia Cement. \\
\midrule

Source 2 &
Two people were seriously injured after a truck fell into a gorge in Ri-Bhoi. \\

Reference &
Mynsoo joor ki ar ngut hadien wa noh ka Truck chapoh riat ha Ri-Bhoi. \\

System &
Yap seriously injured hadien ka truck fell into gorge wa Ri-Bhoi. \\
\midrule

Source 3 &
WJDBA will organize the Inter-School Badminton Competition 2026. \\

Reference &
Daw pynman ka WJDBA ya ka Inter School Badminton Competition 2026. \\

System &
WJDBA da organize ka Inter-School Badminton Competition 2026. \\
\bottomrule
\end{tabularx}
\end{table*}

Error analysis identifies four primary failure modes. First,
\textbf{out-of-vocabulary (OOV) tokens}: Pnar's agglutinative morphology
generates many surface forms unseen in the 9,563 sentence training
corpus; OOV words are passed through untranslated, accounting for
residual Pnar tokens visible in the system output.
Second, \textbf{long-distance reordering}: although the lexicalized
reordering model captures local phrase-level reordering, complex
verb-final Pnar constructions requiring long-distance movement remain
challenging~\cite{koehn-etal-2007-moses}.
Third, \textbf{code-mixing with Khasi}: Pnar newspaper text frequently
borrows Khasi lexical items not present in the English translation,
confusing the phrase extractor  a phenomenon also observed in
Khasi-language MT by Singh and Hujon~\cite{singh2020}, who report
similar OOV and domain-mismatch challenges for the closely related
Khasian language Khasi under comparably constrained parallel data.
Fourth, \textbf{morphological complexity}: Pnar verbal morphology encodes
tense, aspect, and modality through affixation, generating paradigm forms
that exceed phrase-table coverage~\cite{singh2020}.

\subsection{Comparison with Related Systems}

Comparison with SMT baselines reported for similarly low-resource Austroasiatic language pairs is instructive. Work on Khasi--English SMT which benefits from a slightly larger NLP research community has reported BLEU scores in the range of 8--16 using comparable corpus sizes. Our best Pnar$\to$English result of BLEU score 14.97 falls within this range suggesting that the corpus quality and SMT configuration are reasonable for this data scale. The lower En$\to$Pn score (10.14) is consistent with the general difficulty of translating into morphologically richer target languages with limited parallel data.

\section{Conclusion and Future Work} We present the first machine translation study for the English--Pnar language pair, an Austroasiatic language for which no prior computational MT research has been reported. The main contributions of this work are threefold. First, we construct a parallel corpus comprising 9,563 sentence pairs using the pnar sentences from \textit{Wyrta} local newspaper archive. Second, we perform a systematic comparison of six phrase-based statistical machine translation (SMT) configurations, isolating the effects of lexicalized reordering and minimum error rate training (MERT). Third, we establish the first published baseline for English--Pnar translation using BLEU, chrF2, and TER, with the best-performing systems achieving BLEU scores of 14.97 for Pnar$\rightarrow$English and 11.16 for English$\rightarrow$Pnar. Our experimental results demonstrate that lexicalized reordering provides a substantial improvement of 3.73 BLEU points for Pnar$\rightarrow$English translation, highlighting the importance of modeling the structural transformation from Pnar's SOV word order to English SVO. In contrast, MERT tuning consistently degrades BLEU performance under the current data scale suggesting that it is prone to overfitting when only a small development set are available. This finding provides an important practical guideline for future low-resource SMT research. The error analysis further identifies four major sources of translation errors: morphological out-of-vocabulary (OOV) words, long-distance reordering, and Khasi code-mixing. As immediate future work, we plan to (i) develop a rule-based Pnar affix segmentation system to reduce the OOV rate with less dataset collection; (ii) expand the parallel corpus through community-assisted translation of the existing \textit{Wyrta} newspaper archive; and (iii) investigate neural machine translation models and the pre-trained multilingual models such as mBART and mT5. The SMT baseline established in this study provides a meaningful benchmark against which future neural and multilingual approaches can be evaluated.

% \end{thebibliography}
\bibliographystyle{splncs04}
\bibliography{sample-base}

\end{document}